\documentclass{article}

\usepackage[T1]{fontenc}
\usepackage[utf8]{inputenc}
\usepackage{graphicx}
\usepackage{float}
\usepackage{authblk}
\usepackage{amsmath, amssymb}
\usepackage{booktabs}
\usepackage{multirow}
\usepackage{algorithm}
\usepackage{algpseudocode}
\usepackage{subfigure}
\usepackage{hyperref}
\usepackage{xcolor}
\usepackage{url}
\usepackage{geometry}
\usepackage{placeins}

\title{SeamEdit: A Black-Box VLM-Agnostic Pipeline for\\
       Large-Image Semantic Editing}

\author[1]{Xiangyu Lyu}
\author[2]{Dan Lei}

\affil[1]{Technische Universit\"{a}t Darmstadt, Darmstadt, Germany}
\affil[2]{Fine-Arts Educator, Yuncheng Middle School, Yuncheng, Shanxi, China}

\date{June 2026}

\begin{document}

\maketitle

\begin{abstract}
Semantic region editing for large images must satisfy two requirements at the same time: high generative quality and natural integration with surrounding content.
Some related methods rely on white-box models and leave the strong generation capability of closed-source models underexplored.
Directly applying closed-source models to tiled editing, however, introduces several failure modes: semantic deformation, canvas-level alignment drift, and visible seam artifacts.
This paper presents SeamEdit, a training-free and model-agnostic pipeline that treats any VLM with inpainting capability as a black-box oracle.
SeamEdit mitigates these issues through a five-stage post-hoc pipeline: 
overlay-based tile decomposition, black-box VLM inpainting, geometric and color-consistency correction,
seam-risk-based multi-candidate ranking, and dynamic-programming curved seam fusion.
The pipeline reduces seam visibility and supports semantic modification of arbitrary tile regions.
Our code is available at \url{https://github.com/lvxiangyu11/SeamEdit}.
\end{abstract}

\section{Introduction}
\label{sec:intro}

With the development of vision-language models (VLMs) and diffusion-based generative models,
the quality of AI-generated and AI-edited images has continued to improve~\cite{ho2020ddpm, Blattmann_2023_CVPR,podell2023sdxl,brooks2023instructpix2pix}.
However, applying these models to large images still faces practical challenges:
most VLMs operate at fixed input resolutions, which makes tile-based processing necessary for high-resolution content editing.
In practical scenarios such as game art asset production, concept-art iteration, and semantic modification of large maps,
users need to edit local semantic regions in images that exceed the input resolution of a single VLM call,
while preserving natural continuity between the edited region and its surrounding content~\cite{kwon2025mobilepicasso,xie2024rdistitcher,madar2025tiled}.

\textbf{Problem formulation.}
When closed-source black-box VLMs are directly used as inpainting backends for tiled editing,
two types of failure commonly appear.
The first is \textit{alignment drift}:
black-box models usually do not expose pixel-level geometric consistency constraints, and the generated tile may have slight translation, rotation, or scale deviations relative to its original canvas position.
The second is \textit{seam artifacts}:
adjacent tiles generated independently lack cross-tile context, leading to visible discontinuities along tile boundaries.

\textbf{Limitations of existing methods.}
Existing tile-based high-resolution processing methods, such as MultiDiffusion~\cite{bar2023multidiffusion},
MobilePicasso~\cite{kwon2025mobilepicasso}, and HiPrompt~\cite{liu2024hiprompt},
extend generation resolution and improve local consistency by controlling diffusion sampling, latent projection, or patch-wise prompting mechanisms.
These methods require access to intermediate states in the diffusion pipeline, and therefore cannot be directly applied to closed-source VLMs that expose only an API-level input-output interface.
RDIStitcher~\cite{xie2024rdistitcher} introduces a quality assessment metric based on multimodal large models,
but its generation backend is still a self-supervised text-to-image diffusion model.
Therefore, large-image semantic editing with black-box VLMs still requires a processing pipeline that does not rely on model-internal states.

\textbf{This work.}
We propose SeamEdit, a training-free and VLM-agnostic pipeline that treats any VLM with inpainting capability as a black-box oracle and introduces dedicated mechanisms for the failure modes described above.
The main contributions of this paper are as follows:

\begin{enumerate}
    \item \textbf{Overlay-based black-box tiled editing}:
    an overlay-tile decomposition and masking strategy that enables large-image semantic editing
    through black-box VLM inpainting, while providing boundary context and overlap buffers for
    subsequent correction and fusion.

    \item \textbf{Post-hoc candidate correction and ranking}:
    a candidate-level processing module that combines Grid-SIFT geometric alignment,
    local color-consistency correction, and seam-risk-based quality scoring to select reliable
    VLM-generated tile results without accessing model-internal states.

    \item \textbf{Seam-aware integration of selected tile candidates}: a post-hoc fusion stage that adapts content-adaptive seam search and feathered blending to the black-box VLM tiled-editing setting, merging selected candidates back into the global canvas while reducing visible boundary artifacts.
\end{enumerate}

The rest of this paper is organized as follows.
Section~\ref{sec:related} reviews related work.
Section~\ref{sec:method} describes each module of the pipeline.
Section~\ref{sec:experiments} reports experimental results.
Section~\ref{sec:discussion} discusses limitations and application scenarios.
Section~\ref{sec:conclusion} concludes the paper.

\section{Related Work}
\label{sec:related}

\subsection{Tile-Based High-Resolution Image Processing}

To overcome the resolution limit of a single model call,
researchers have proposed various tile-based methods for high-resolution image generation and editing.
Diffusion models and their latent-space variants provide widely used generative backends for high-quality image synthesis~\cite{ho2020ddpm,rombach2022high,podell2023sdxl}.
MultiDiffusion~\cite{bar2023multidiffusion} achieves training-free large-image generation by fusing denoising directions over overlapping crops, but repeated texture patterns may still occur.
MobilePicasso~\cite{kwon2025mobilepicasso} decomposes high-resolution editing into standard-resolution editing, latent projection, and context-aware tiled upsampling for improved mobile efficiency.
HiPrompt~\cite{liu2024hiprompt} uses a multimodal large model to generate patch-wise local descriptions and combines them with a global prompt to improve semantic consistency in high-resolution generation.
Tiled Diffusion~\cite{madar2025tiled} generates seamlessly tileable images under multiple tile-connection constraints.
These methods all rely on access to internal diffusion-model states, such as latent variables, denoising directions, or attention maps, whereas closed-source VLMs expose only input-output interfaces; therefore, they cannot be directly used for black-box VLM tiled editing.

\subsection{Diffusion-Based Image Stitching}

Recent work in image stitching focuses on removing seams and tonal inconsistencies that arise when multiple images are merged.
Classical panorama stitching typically relies on local invariant features for image matching and registration~\cite{brown2007stitching},
and uses multi-resolution blending or gradient-domain editing to smooth boundary transitions~\cite{burt1983multiresolution,perez2003poisson}.
RDIStitcher~\cite{xie2024rdistitcher} reformulates image fusion and rectangling as a reference-based inpainting problem and introduces a multimodal-large-model-based quality assessment metric,
but its generative backend remains a self-supervised fine-tuned text-to-image diffusion model that requires white-box training access.
Seamless-Through-Breaking~\cite{chen2024seamless} addresses geometric alignment for panoramic stitching from multiple photographs,
which differs from the single-large-image tiled semantic editing setting considered in this paper.
This class of methods targets different usage scenarios and does not directly cover alignment drift in black-box VLM tiled editing.

\subsection{VLM-Based Image Editing}

Recent work has explored the use of VLMs for semantic image editing.
In diffusion-based image editing, InstructPix2Pix~\cite{brooks2023instructpix2pix} and MagicBrush~\cite{zhang2023magicbrush} focus on natural-language-instruction-driven real-image editing,
while Blended Latent Diffusion~\cite{avrahami2022blended} restricts editing to user-specified local regions.
UniEdit-I~\cite{bai2025uniedit} proposes a training-free closed-loop editing framework in a unified VLM semantic latent space,
but it focuses on single-image closed-loop editing and does not address large-image tiling or seam fusion.
The above work does not address alignment drift and seam artifacts that arise when black-box VLMs are applied to large-image tiled editing.
This paper focuses on this more specific tiled editing setting.

\section{Method}
\label{sec:method}

\subsection{Pipeline Overview}
\label{sec:overview}

\begin{figure}[H]
    \centering
    \includegraphics[width=\linewidth,height=0.74\textheight,keepaspectratio]{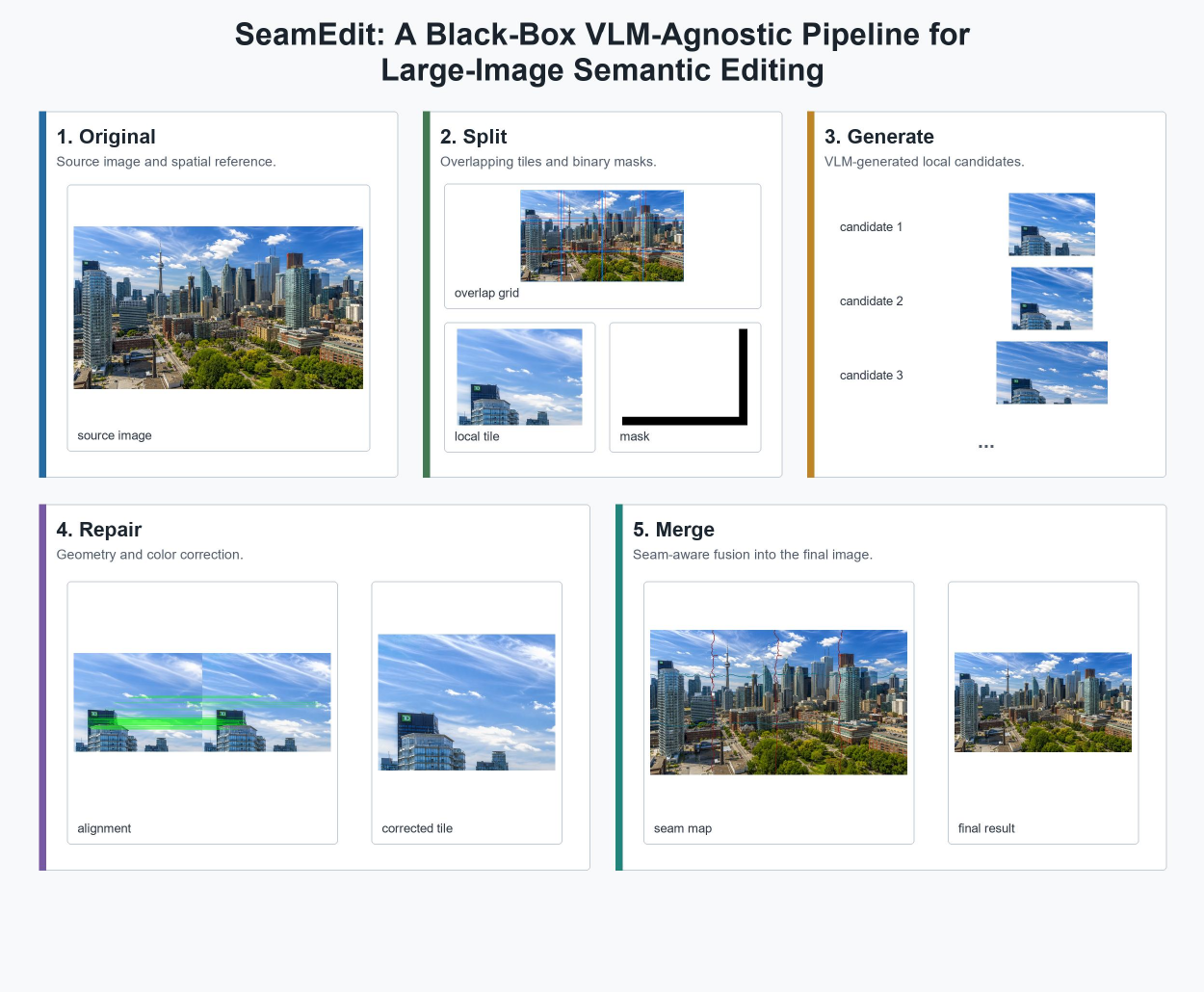}
    \caption{Overview of the SeamEdit pipeline. The five stages are tile decomposition, VLM inpainting, geometric and color correction, multi-candidate ranking, and dynamic-programming curved seam fusion.}
    \label{fig:pipeline}
\end{figure}

Figure~\ref{fig:pipeline} shows the five-stage pipeline of SeamEdit.
Given a large image $\mathcal{I} \in \mathbb{R}^{H \times W \times 3}$,
the user selects one or more tile regions for semantic editing.
The pipeline proceeds through the following stages:
(1) overlay-based tile decomposition and mask generation;
(2) VLM inpainting for each tile, where each tile produces $K$ candidate results;
(3) geometric and color-consistency correction;
(4) seam risk measurement and multi-candidate ranking;
(5) dynamic-programming curved seam fusion, which stitches the selected candidates back into the global image.

\subsection{Tile Decomposition with Overlay Masks}
\label{sec:overlay_decomposition}
\begin{figure}[t]
    \centering
    \includegraphics[width=\linewidth]{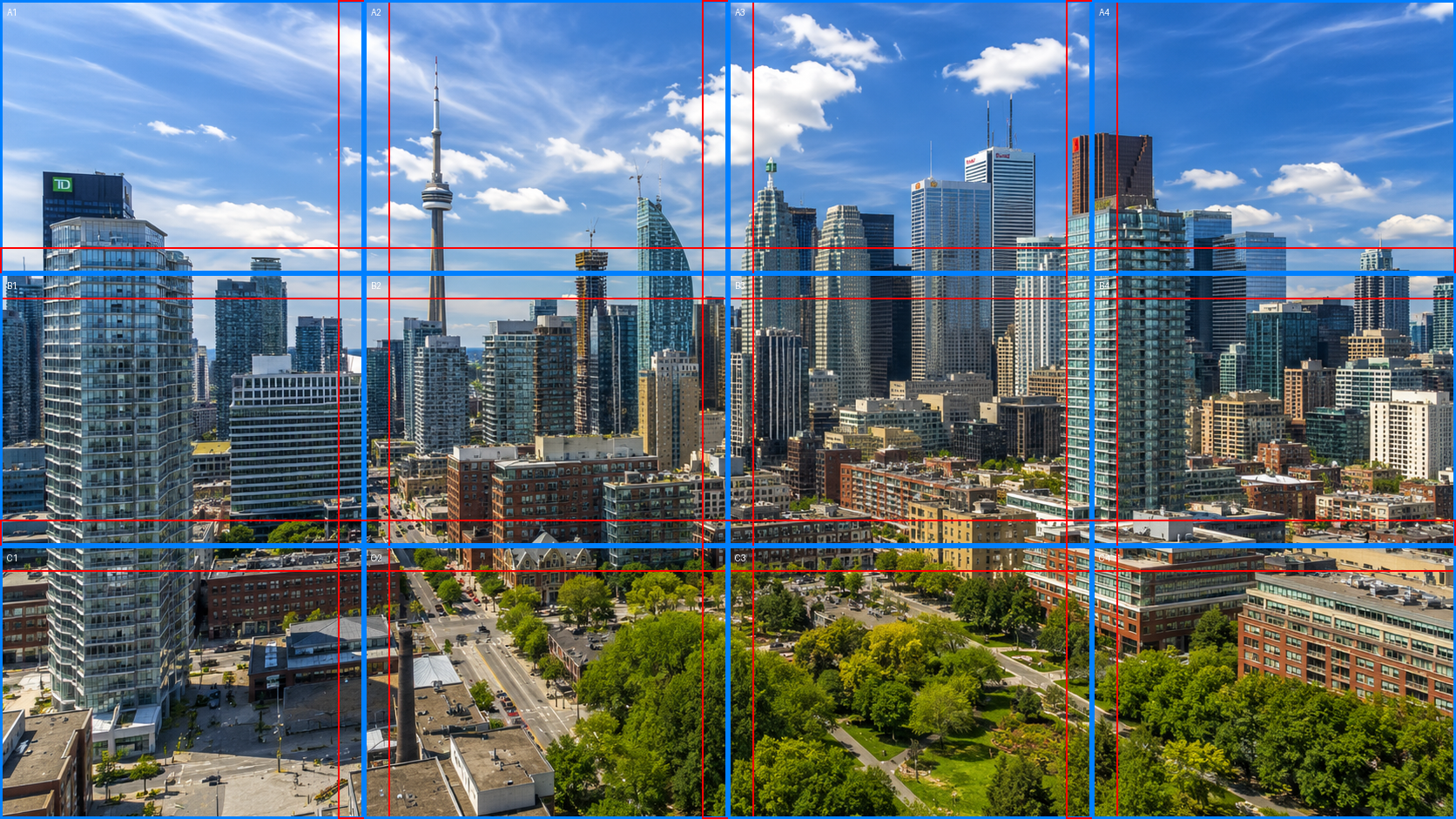}
    \caption{
    Overlay-based tile decomposition. The thick blue lines indicate the core region $C_{r,c}$,
    and the thin red frame indicates the tile region $T_{r,c}$ obtained by expanding the core by $\delta$ pixels in all directions.
    The expanded region provides boundary context for black-box VLM inpainting and an overlap buffer for subsequent seam fusion.
    }
    \label{fig:overlay_decomposition}
\end{figure}

Given an input large image $I \in \mathbb{R}^{H \times W \times 3}$, SeamEdit first divides the image into an $R \times C$ regular grid.
For the tile at row $r$ and column $c$, its core region $C_{r,c}$ is defined as the independent image region assigned to this tile during final stitching:
\[
C_{r,c} = [x_c^0, x_c^1) \times [y_r^0, y_r^1),
\]
where
\[
x_c^0 = \text{round}(cW/C), \quad x_c^1 = \text{round}((c+1)W/C),
\]
\[
y_r^0 = \text{round}(rH/R), \quad y_r^1 = \text{round}((r+1)H/R).
\]

To provide local boundary context to the black-box VLM, SeamEdit expands each core region by an overlay width $\delta$ on all four sides.
The actual tile region $T_{r,c}$ sent to the VLM is defined as
\[
T_{r,c} = [\max(0, x_c^0 - \delta), \min(W, x_c^1 + \delta)) \times [\max(0, y_r^0 - \delta), \min(H, y_r^1 + \delta)).
\]

Here, $\delta$ denotes the overlay width.
Tiles located on image boundaries are clipped to the original image canvas, so corner and edge tiles naturally have smaller overlay regions.
In the implementation, the program first computes the core rectangle from the regular grid, expands it with the overlay, and then constrains the tile rectangle to the image range using a \texttt{clamp} operation.

After cropping the tile image, SeamEdit converts the core-region coordinates into the local tile coordinate system.
Let $(x_T^0, y_T^0)$ denote the top-left coordinate of $T_{r,c}$.
The local core region is
\[
\tilde{C}_{r,c} = C_{r,c} - (x_T^0, y_T^0).
\]

Based on this local core region, the system generates a binary inpainting mask $M_{r,c}$ for each tile:
\[
M_{r,c}(p) = \begin{cases} 255, & p \in \tilde{C}_{r,c}, \\ 0, & p \in T_{r,c} \setminus \tilde{C}_{r,c}. \end{cases}
\]

The tile image and its corresponding mask are then resized according to the target VLM input size.
The RGB image uses Lanczos interpolation to preserve more texture details, while the binary mask uses nearest-neighbor interpolation to keep a sharp discrete boundary.
The system also records spatial metadata including the core region, overlay region, tile size, scale ratio, and four directional overlay subregions at both the original resolution and the output resolution.
These metadata are reused in the subsequent alignment correction, candidate ranking, and final fusion stages.

This decomposition plays three roles.
First, it provides boundary context to the VLM so that generated content can naturally continue the surrounding image.
Second, it provides feature-matching regions for later alignment correction, improving geometric consistency.
Third, it provides an operational buffer for seam processing so that cuts can be placed away from visually salient core content.

\subsection{Tile Repainting and Multi-Candidate Generation}

After obtaining the overlay tile $T_{r,c}$ and its binary mask $M_{r,c}$, SeamEdit feeds each tile into a vision-language model with local inpainting capability.
This paper treats the model as a black-box generator and assumes only that it can receive an image, a mask, and a text editing instruction, and then return an inpainted image.
The entire workflow depends only on the model's input-output interface, making it compatible with different closed-source or open-source VLM backends.

For the $(r,c)$-th tile, the input consists of three components: the local image $T_{r,c}$, the inpainting mask $M_{r,c}$, and the user-provided semantic editing instruction $q$.
The white region in the mask corresponds to the core region to be repainted, while the black region corresponds to the overlay area preserved as visual context.
The VLM semantically inpaints the core region according to the structure, color, and texture information in the surrounding overlay area:
\[
\hat{T}_{r,c} = \mathcal{G}(T_{r,c}, M_{r,c}, q),
\]
where $\mathcal{G}$ denotes the black-box VLM inpainting function and $\hat{T}_{r,c}$ denotes the generated local tile result.

Because the generation process of a black-box VLM is stochastic, the same input may produce different local structures, texture details, and boundary transitions.
To increase the selection space in subsequent stages, SeamEdit repeatedly generates $K$ candidate results for each tile:
\[
\mathcal{Y}_{r,c} = \{ \hat{T}_{r,c}^1, \hat{T}_{r,c}^2, \dots, \hat{T}_{r,c}^K \},
\]
where
\[
\hat{T}_{r,c}^k = \mathcal{G}(T_{r,c}, M_{r,c}, q; \xi_k),
\]
and $\xi_k$ denotes the stochastic sampling state of the $k$-th generation process.
All candidates share the same tile input, mask constraint, and text instruction; their differences come only from the model's internal sampling variation.

This multi-candidate mechanism separates generation from selection.
The generation stage focuses on obtaining diverse local inpainting results, while later stages correct and rank candidates according to geometric consistency, boundary consistency, and seam risk.
In this way, SeamEdit uses the generative capability of black-box VLMs for local semantic editing of large images without accessing model parameters, attention maps, or intermediate diffusion states.

\subsection{Geometric and Color-Consistency Correction}
\label{sec:geo_photo_correction}

During local inpainting, a black-box VLM may alter the pixel-level geometry of a tile, causing slight translation, rotation, or scale deviations of the generated candidate relative to the reference tile.
At the same time, the overall tone, brightness, and local low-frequency color distribution of the generated candidate may differ from the original image context.
These two types of deviations appear as structural misalignment and color discontinuity when the tile is fused back into the global image.
Therefore, before candidate ranking and final fusion, SeamEdit applies geometric alignment and color-consistency correction to each generated candidate.

Let $T_{r,c}$ be the reference tile at position $(r,c)$, and let $\hat{T}^{k}_{r,c}$ be the $k$-th VLM-generated candidate for this tile.
The goal of this stage is to map the generated candidate back to the reference tile coordinate system and obtain a color-corrected candidate:

\begin{equation}
\bar{T}^{k}_{r,c}
=
\mathcal{P}
\left(
\mathcal{W}\left(\hat{T}^{k}_{r,c}; A^{k}_{r,c}\right),
T_{r,c}
\right),
\end{equation}

where $A^{k}_{r,c} \in \mathbb{R}^{2 \times 3}$ denotes a two-dimensional affine transform from the generated-candidate coordinate system to the reference-tile coordinate system, $\mathcal{W}(\cdot)$ denotes affine resampling, and $\mathcal{P}(\cdot)$ denotes local color-consistency correction.
After this step, the candidate result shares the same local reference with the reference tile in both geometric coordinates and color distribution.

\subsubsection{Grid-SIFT Geometric Alignment}

Local feature detection can concentrate in a few high-texture regions, causing the estimated geometric transformation to depend too heavily on local structures.
To improve feature coverage over the tile space, SeamEdit uses grid-based SIFT feature detection.
Given a matching image $I \in \mathbb{R}^{h \times w \times 3}$, the image is divided into $G_r \times G_c$ rectangular subregions.
The $(i,j)$-th subregion is defined as

\begin{equation}
\Omega_{i,j}
=
[x^0_j, x^1_j)
\times
[y^0_i, y^1_i),
\end{equation}

where

\begin{equation}
x^0_j = \mathrm{round}(jw/G_c), 
\quad
x^1_j = \mathrm{round}((j+1)w/G_c),
\end{equation}

\begin{equation}
y^0_i = \mathrm{round}(ih/G_r), 
\quad
y^1_i = \mathrm{round}((i+1)h/G_r).
\end{equation}

Within each subregion $\Omega_{i,j}$, SeamEdit independently extracts local SIFT features.
The detected local keypoint coordinate $\tilde{p}$ lies in the subregion coordinate system and is therefore mapped back to the full tile coordinate system:

\begin{equation}
p = \tilde{p} + (x^0_j, y^0_i).
\end{equation}

The keypoints and descriptors from all subregions jointly form the feature set of the whole tile:

\begin{equation}
\mathcal{F}(I)
=
\bigcup_{i=0}^{G_r-1}
\bigcup_{j=0}^{G_c-1}
\mathcal{D}(I|_{\Omega_{i,j}}),
\end{equation}

where $\mathcal{D}(\cdot)$ denotes the local SIFT feature extraction operator.
This grid-based strategy allows local textures, edge structures, and region boundaries at different spatial locations to participate in transformation estimation, improving the global reliability of geometric alignment.

For the reference tile $T_{r,c}$ and the generated candidate $\hat{T}^{k}_{r,c}$, SeamEdit extracts feature sets $\mathcal{F}(T_{r,c})$ and $\mathcal{F}(\hat{T}^{k}_{r,c})$ respectively, and obtains candidate correspondences through descriptor matching.
Let $p_i$ be a feature point in the reference tile and $q_i$ be its matched feature point in the generated candidate.
The inlier set obtained by robust estimation is denoted as $\mathcal{I}$.
The affine transformation is obtained by minimizing the inlier reprojection error:

\begin{equation}
A^{k}_{r,c}
=
\arg\min_A
\sum_{(p_i,q_i)\in \mathcal{I}}
\left\|
p_i - A q_i
\right\|_2^2.
\end{equation}

This objective maps point $q_i$ in the generated candidate to the corresponding location $p_i$ in the reference tile.
Thus, $A^{k}_{r,c}$ is defined as the transform from the candidate coordinate system to the reference coordinate system.
The resampling operator $\mathcal{W}(\cdot;\,A)$ is implemented through backward sampling:
for each output pixel $p$ on the reference canvas, the value is sampled from $A^{-1}p$ in the candidate image,
thereby mapping the candidate result onto the reference tile canvas:

\begin{equation}
\tilde{T}^{k}_{r,c}
=
\mathcal{W}\left(\hat{T}^{k}_{r,c}; A^{k}_{r,c}\right).
\end{equation}

\begin{figure}[t]
    \centering
    \includegraphics[width=\linewidth]{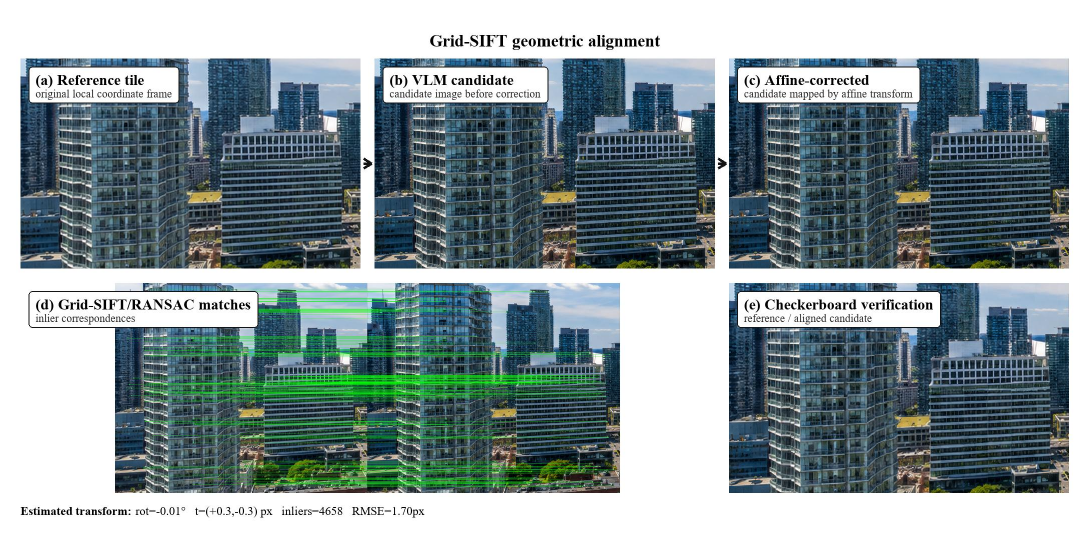}
    \caption{
    Example of Grid-SIFT geometric alignment.
    The reference tile and the VLM-generated candidate are matched using spatially uniform SIFT features.
    RANSAC estimates the affine transformation and maps the generated candidate back to the reference tile coordinate system.
    The checkerboard visualization shows the local alignment after correction.
    }
    \label{fig:grid_sift_alignment}
\end{figure}

To control abnormal deformation caused by erroneous matches, SeamEdit applies geometric validity constraints to the estimated result.
A valid transformation must satisfy sufficient inlier count, reasonable spatial coverage of inliers, low reprojection error, and bounded translation, rotation, scale change, and shear.
This constraint is written as

\begin{equation}
A^{k}_{r,c} \in \mathcal{A},
\end{equation}

where

\begin{equation}
\begin{aligned}
\mathcal{A}
=
\{ A \mid\;&
N_{\mathrm{in}} \ge \tau_{\mathrm{in}},
\quad
C_{\mathrm{cov}} \ge \tau_{\mathrm{cov}},
\quad
E_{\mathrm{rmse}} \le \tau_{\mathrm{rmse}}, \\
&
|t_x| \le \tau_x,
\quad
|t_y| \le \tau_y,
\quad
|\theta| \le \tau_\theta, \\
&
|s_x-s_x^\ast| \le \tau_s,
\quad
|s_y-s_y^\ast| \le \tau_s,
\quad
|h| \le \tau_h
\}.
\end{aligned}
\end{equation}

Here, $N_{\mathrm{in}}$ denotes the number of inliers in robust estimation, $C_{\mathrm{cov}}$ denotes the number of spatial regions covered by inliers, $E_{\mathrm{rmse}}$ denotes the inlier reprojection error, $(t_x,t_y)$ denotes translation, $\theta$ denotes the rotation angle, $(s_x,s_y)$ denotes scale, $(s_x^\ast,s_y^\ast)$ denotes the expected scale given by the input-output size relation, and $h$ denotes shear.
Low-confidence estimates use the identity transform:

\begin{equation}
A^{k}_{r,c}=I_{2\times3}.
\end{equation}

Thus, SeamEdit uses affine correction only when sufficient geometric evidence is available, reducing the risk of candidate-image deformation caused by local mismatches.

\subsubsection{Local Color-Consistency Correction}

Geometric alignment maps the generated candidate back into the reference tile coordinate system, but the candidate may still contain global tonal shifts and low-frequency color differences.
To make the candidate consistent with the surrounding context in illumination and color distribution, SeamEdit performs local color normalization within the valid candidate region.

Let $\tilde{T}^{k}_{r,c}$ be the geometrically aligned candidate, and let
$V^{k}_{r,c} = \{p \mid \mathcal{W}(\hat{T}^{k}_{r,c};\,A^{k}_{r,c})(p) \text{ is valid}\}$
denote the set of non-hole pixels after affine resampling, namely the pixels in the output domain that can be covered by the candidate image.
SeamEdit computes per-channel robust statistics of the reference tile and the candidate tile within $V^{k}_{r,c}$:

\begin{equation}
\mu_T,\sigma_T
=
\mathrm{Stat}
\left(
T_{r,c}, V^{k}_{r,c}
\right),
\end{equation}

\begin{equation}
\mu_{\tilde{T}},\sigma_{\tilde{T}}
=
\mathrm{Stat}
\left(
\tilde{T}^{k}_{r,c}, V^{k}_{r,c}
\right),
\end{equation}

where $\mu$ and $\sigma$ denote the RGB channel mean and standard deviation, respectively, and $\mathrm{Stat}(\cdot)$ denotes robust channel statistics over pixels inside the mask region.
Based on these statistics, the per-channel color gain and bias are defined as

\begin{equation}
g =
\mathrm{clip}
\left(
\frac{\sigma_T}{\sigma_{\tilde{T}}},
g_{\min},
g_{\max}
\right),
\end{equation}

\begin{equation}
b =
\mathrm{clip}
\left(
\mu_T - g \odot \mu_{\tilde{T}},
b_{\min},
b_{\max}
\right),
\end{equation}

where $\odot$ denotes per-channel multiplication, and $\mathrm{clip}(\cdot)$ limits the magnitude of the color transform.
The color-matched candidate is

\begin{equation}
\tilde{T}^{k,\mathrm{col}}_{r,c}(p)
=
(1-\eta)\tilde{T}^{k}_{r,c}(p)
+
\eta
\left(
g \odot \tilde{T}^{k}_{r,c}(p) + b
\right),
\quad p \in V^{k}_{r,c},
\label{eq:color-match-blend}
\end{equation}

where $\eta$ controls the strength of color correction.
This linear channel transform moves the candidate region's mean and contrast toward those of the reference tile, while magnitude limits preserve the generated result's texture details.

Using only global channel statistics may be insufficient for slowly varying low-frequency color differences.
SeamEdit therefore further introduces low-frequency color compensation.
Let $\mathcal{G}_{\rho}(\cdot)$ denote Gaussian low-pass filtering at scale $\rho$.
The low-frequency difference between the reference tile and the current candidate is

\begin{equation}
\Delta^{k}_{r,c}
=
\mathrm{clip}
\left(
\mathcal{G}_{\rho}(T_{r,c})
-
\mathcal{G}_{\rho}(\tilde{T}^{k,\mathrm{col}}_{r,c}),
-\epsilon,
\epsilon
\right).
\end{equation}

The final color-corrected candidate is defined as

\begin{equation}
\bar{T}^{k}_{r,c}(p)
=
\tilde{T}^{k,\mathrm{col}}_{r,c}(p)
+
\gamma \Delta^{k}_{r,c}(p),
\quad p \in V^{k}_{r,c},
\label{eq:lowfreq-color-correction}
\end{equation}

where $\gamma$ denotes the low-frequency compensation strength, and $\epsilon$ limits the magnitude of a single low-frequency correction.
For regions outside $V^{k}_{r,c}$, the candidate canvas keeps the reference tile content:

\begin{equation}
\bar{T}^{k}_{r,c}(p)=T_{r,c}(p),
\quad p \notin V^{k}_{r,c}.
\end{equation}

After geometric and color-consistency correction, each VLM candidate is represented as a corrected result $\bar{T}^{k}_{r,c}$ in the reference tile coordinate system.
The subsequent seam risk metric, multi-candidate ranking, and dynamic seam fusion are all based on this corrected candidate, reducing the influence of canvas drift and color drift in black-box VLM outputs on the final large-image editing result.

\begin{figure}[t]
    \centering
    \includegraphics[width=\linewidth]{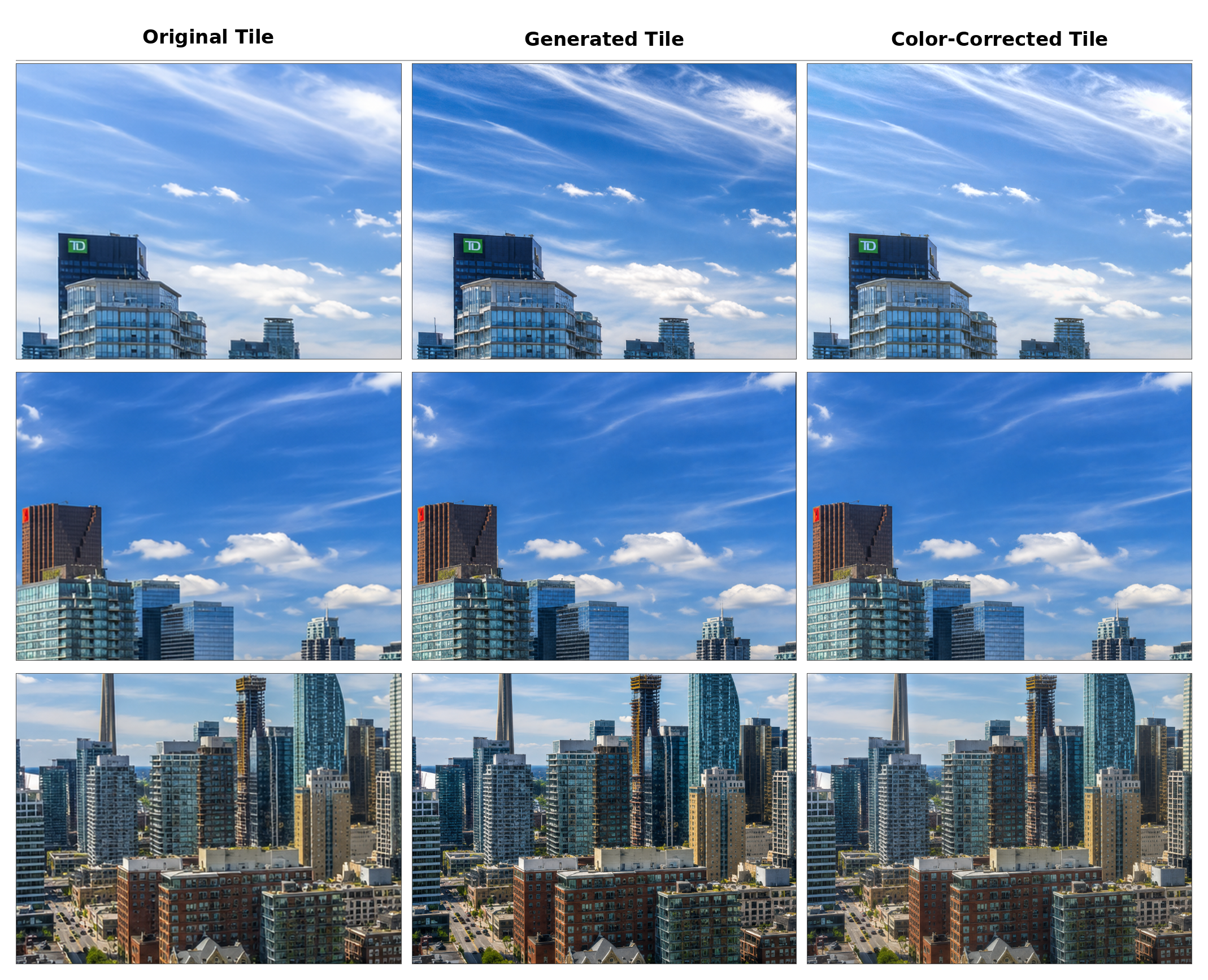}
    \caption{
    Example of local color-consistency correction.
    Each row shows the original tile, the VLM-generated tile, and the color-corrected tile.
    Color correction reduces brightness and tonal shifts in the generated result while preserving local structural details.
    }
    \label{fig:color_correction_example}
\end{figure}

\subsubsection{Seam Risk Scoring and Candidate Ranking}
\label{sec:seam-risk}

After geometric alignment and color-consistency correction, SeamEdit scores each tile candidate before final stitching.
This score estimates the risk that a candidate image will produce visible seams in the overlay region and selects the best candidate among multiple generations for the same tile.

Let $\bar{T}_{r,c}^{k}$ denote the $k$-th corrected candidate of the $(r,c)$-th tile, and let $T_{r,c}$ denote the corresponding reference tile.
The local tile region is denoted as $\Omega_{r,c}$, and the core repainting region is denoted as $C_{r,c}$.
The overlay evaluation region is defined as
\begin{equation}
B_{r,c}=\Omega_{r,c}\setminus C_{r,c}.
\end{equation}

For each valid seam direction $s\in\mathcal{S}_{r,c}$, where $\mathcal{S}_{r,c}$ may contain the left, right, top, and bottom directions, the corresponding overlay boundary band is denoted as $B_{r,c}^{s}\subset B_{r,c}$.
The color discontinuity error of candidate $k$ in direction $s$ is defined as
\begin{equation}
E_{\mathrm{col}}^{k,s}
=
\frac{1}{3|B_{r,c}^{s}|}
\sum_{p\in B_{r,c}^{s}}
\left\|
\bar{T}_{r,c}^{k}(p)-T_{r,c}(p)
\right\|_{1}.
\label{eq:seam-color-error}
\end{equation}

To further characterize local structural differences, this paper also computes the gradient residual:
\begin{equation}
E_{\mathrm{grad}}^{k,s}
=
\frac{1}{3|B_{r,c}^{s}|}
\sum_{p\in B_{r,c}^{s}}
\left\|
\nabla \bar{T}_{r,c}^{k}(p)-\nabla T_{r,c}(p)
\right\|_{1}.
\label{eq:seam-gradient-error}
\end{equation}

The seam risk of candidate $k$ is defined as the maximum boundary risk among all valid directions:
\begin{equation}
R_{r,c}^{k}
=
\max_{s\in\mathcal{S}_{r,c}}
\left(
\beta_{\mathrm{col}}E_{\mathrm{col}}^{k,s}
+
\beta_{\mathrm{grad}}E_{\mathrm{grad}}^{k,s}
\right),
\label{eq:seam-risk-score}
\end{equation}
where $\beta_{\mathrm{col}}$ and $\beta_{\mathrm{grad}}$ control the relative weights of color difference and gradient difference, respectively.
A higher risk indicates that the candidate image is more likely to produce visible discontinuities around seam regions.

Candidate ranking also considers geometric alignment quality, valid pixel coverage, and overall color consistency.
Let $E_{\mathrm{geo}}^{k}$ denote the geometric penalty derived from Grid-SIFT alignment residuals, inlier ratio, and scale consistency; let $E_{\mathrm{valid}}^{k}$ denote the invalid-pixel penalty after affine resampling; and let $E_{\mathrm{color}}^{k}$ denote the overall color difference between the candidate and the reference tile.
The final candidate quality score is defined as
\begin{equation}
Q_{r,c}^{k}
=
\alpha_{\mathrm{psnr}}P_{r,c}^{k}
+
\alpha_{\mathrm{ssim}}U_{r,c}^{k}
-
\alpha_{\mathrm{seam}}R_{r,c}^{k}
-
\alpha_{\mathrm{geo}}E_{\mathrm{geo}}^{k}
-
\alpha_{\mathrm{valid}}E_{\mathrm{valid}}^{k}
-
\alpha_{\mathrm{color}}E_{\mathrm{color}}^{k},
\label{eq:candidate-score}
\end{equation}
where $P_{r,c}^{k}$ and $U_{r,c}^{k}$ denote the PSNR term and SSIM term in the overlay region, respectively.
The final selected candidate is
\begin{equation}
k^{\ast}
=
\arg\max_{k} Q_{r,c}^{k}.
\label{eq:candidate-selection}
\end{equation}

Thus, \textsc{SeamRiskScore} in Algorithm~\ref{alg:seamedit} corresponds to Eq.~\eqref{eq:seam-risk-score}, and automatic multi-candidate selection corresponds to Eq.~\eqref{eq:candidate-score} and Eq.~\eqref{eq:candidate-selection}.
This process preferentially selects candidates whose overlay regions are more consistent with the original context, whose geometric alignment is more reliable, and whose valid pixel coverage is more complete.

\subsection{Dynamic-Programming Curved Seam Fusion}
\label{sec:dp_seam_fusion}

After candidate ranking in Section~\ref{sec:geo_photo_correction}, each tile obtains an optimal candidate $T^{\ast}_{r,c}$.
This section writes all optimal candidates back into the global image canvas.
Direct fusion along regular grid boundaries produces obvious straight seams, so SeamEdit searches for a low-cost curved seam inside the overlap between adjacent regions and performs smooth alpha blending on the two sides of the seam.

\begin{figure}[t] \centering \includegraphics[ width=\linewidth, keepaspectratio ]{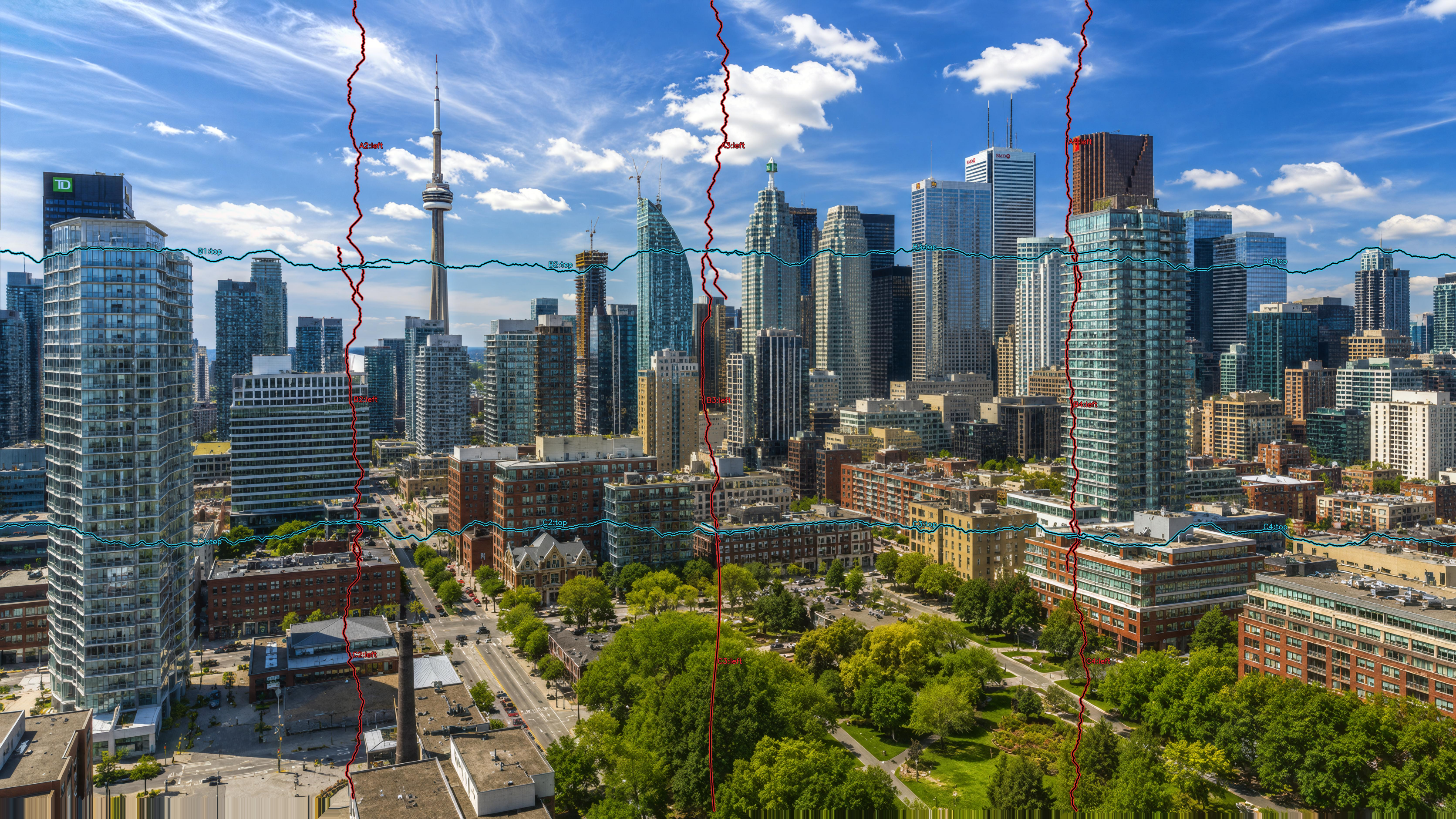} \caption{ Visualization of dynamic-programming curved seam fusion. The figure shows the seam paths in the final fused image. SeamEdit searches for low-cost curved seams within the overlap regions between adjacent tiles and combines them with feathered alpha blending to reduce visible discontinuities caused by regular grid boundaries. } \label{fig:best-merged-seams} \end{figure}

Figure~\ref{fig:best-merged-seams} visualizes the distribution of curved seams in the final fused result. These seams lie within the overlap regions between adjacent tiles and pass through locations with lower color and structural differences, thereby reducing visible discontinuities caused by regular grid boundaries. 

Let the current fused global canvas be $I_{\mathrm{mos}}$, and let its written region be $M_{\mathrm{mos}}$.
According to the spatial metadata recorded in Section~\ref{sec:overlay_decomposition}, the optimal candidate $T^{\ast}_{r,c}$ is placed into the global coordinate system, producing a global candidate image $P_{r,c}$ with valid region $V_{r,c}$.
The overlap between the current tile and the existing canvas is defined as

\begin{equation}
O_{r,c} = V_{r,c} \cap M_{\mathrm{mos}}.
\end{equation}

If $O_{r,c}$ is empty, the candidate region is directly written to the canvas.
Otherwise, SeamEdit computes a curved seam inside this region and uses blending weights to control the transition between the existing canvas and the current candidate.

Inspired by minimum-energy seam search in seam carving~\cite{Avidan2007seam}, SeamEdit formulates seam selection in the overlap region as a one-dimensional dynamic programming problem.

\paragraph{Seam cost.}
For a pixel $p$ inside the overlap region, the seam cost is composed of color difference and structural difference:

\begin{equation}
\mathcal{C}(p)
=
\bigl\|P_{r,c}(p)-I_{\mathrm{mos}}(p)\bigr\|_1
+
\lambda
\bigl\|\nabla P_{r,c}(p)-\nabla I_{\mathrm{mos}}(p)\bigr\|_1,
\quad p \in O_{r,c},
\end{equation}

where $\nabla$ denotes the image gradient operator, and $\lambda$ controls the weight of the structural-difference term.
The color term encourages the seam to pass through regions with similar colors, while the gradient term encourages the seam to avoid salient structural boundaries.

\paragraph{Dynamic-programming curved seam.}
For a vertical overlap region, SeamEdit searches for a top-to-bottom curved seam
$\Gamma = \{(x(y),\,y)\}_{y=0}^{h-1}$,
where $h$ is the height of the overlap region and $x(y)$ denotes the horizontal seam coordinate at row $y$.
The horizontal displacement between adjacent rows satisfies
$x(y)-x(y-1)\in\{-1,0,1\}$.
Let $D(y,x)$ denote the minimum cumulative cost from the first row to position $(x,y)$.
The recurrence relation is

\begin{equation}
D(y,x)
=
\mathcal{C}(y,x)
+
\min\bigl\{D(y-1,x-1),\;D(y-1,x),\;D(y-1,x+1)\bigr\}.
\end{equation}

The first row is initialized as $D(0,x)=\mathcal{C}(0,x)$.
The position with the minimum cumulative cost in the last row is selected as the endpoint, and the optimal path $\Gamma^{\ast}$ is obtained by following the backtracking pointers.
A horizontal overlap region is solved symmetrically in the transposed coordinate system.

\paragraph{Feathered alpha blending.}
After obtaining the optimal seam, SeamEdit constructs smooth alpha weights centered at $\Gamma^{\ast}$.
For a vertical seam, let the seam position at row $y$ be $x^{\ast}(y)$.
The blending weight of the candidate image is defined as

\begin{equation}
\alpha(x,y)
=
\mathrm{clip}\!\left(\frac{x - x^{\ast}(y) + f}{2f},\;0,\;1\right),
\end{equation}

where $f$ is the feathering width.
The side with $\alpha=0$ keeps the existing canvas $I_{\mathrm{mos}}$, while the side with $\alpha=1$ uses the current candidate $P_{r,c}$.
The final fused result in the overlap region is

\begin{equation}
I_{\mathrm{mos}}^{\mathrm{new}}(p)
=
(1-\alpha(p))\,I_{\mathrm{mos}}(p)
+\alpha(p)\,P_{r,c}(p),
\quad p \in O_{r,c}.
\end{equation}

The candidate content is directly written in non-overlap regions, and $M_{\mathrm{mos}}$ is updated.
After all tiles are processed, the final output is $I_{\mathrm{out}}=I_{\mathrm{mos}}$.
This process replaces regular grid boundaries with content-adaptive low-cost curved paths, making seam positions pass through regions with smaller color and structural differences whenever possible; feathered blending further reduces visible discontinuities caused by hard boundaries.

Algorithm~\ref{alg:seamedit} summarizes the complete workflow.
\begin{algorithm}[H]
\caption{Overall SeamEdit Pipeline}
\label{alg:seamedit}
\begin{algorithmic}[1]
\Require Input image $I$, editing instruction $q$, grid size $R \times C$, overlay width $\delta$, number of candidates $K$
\Ensure Edited large image $I_{\mathrm{out}}$

\State $\{T_{r,c}, M_{r,c}, \Omega_{r,c}\}_{r=0,c=0}^{R-1,C-1}
\gets \textsc{TileDecompose}(I, R, C, \delta)$

\For{$r = 0$ to $R-1$}
    \For{$c = 0$ to $C-1$}
        \State $\mathcal{B}_{r,c} \gets \emptyset$

        \For{$k = 1$ to $K$}
            \State $\hat{T}^{k}_{r,c}
            \gets
            \mathcal{G}(T_{r,c}, M_{r,c}, q; \xi_k)$

            \State $A^{k}_{r,c}
            \gets
            \textsc{GridSIFTAlign}(\hat{T}^{k}_{r,c}, T_{r,c})$

            \State $\tilde{T}^{k}_{r,c}
            \gets
            \mathcal{W}(\hat{T}^{k}_{r,c}; A^{k}_{r,c})$

            \State $\bar{T}^{k}_{r,c}
            \gets
            \textsc{ColorNormalize}(\tilde{T}^{k}_{r,c}, T_{r,c})$

            \State $\mathrm{Score}^{k}_{r,c}
            \gets
            \textsc{SeamRiskScore}(\bar{T}^{k}_{r,c}, T_{r,c}, M_{r,c}, A^{k}_{r,c})$

            \State $\mathcal{B}_{r,c}
            \gets
            \mathcal{B}_{r,c} \cup \{\bar{T}^{k}_{r,c}\}$
        \EndFor

        \State $k^{\ast}
        \gets
        \arg\max_{k \in \{1,\ldots,K\}}
        \mathrm{Score}^{k}_{r,c}$

        \State $T^{\ast}_{r,c}
        \gets
        \bar{T}^{k^{\ast}}_{r,c}$
    \EndFor
\EndFor

\State Initialize the global canvas $I_{\mathrm{mos}}$ and the written-region mask $M_{\mathrm{mos}}$

\For{$r = 0$ to $R-1$}
    \For{$c = 0$ to $C-1$}
        \State $(P_{r,c}, V_{r,c})
        \gets
        \textsc{PlaceOnCanvas}(T^{\ast}_{r,c}, \Omega_{r,c})$

        \State $O_{r,c}
        \gets
        V_{r,c} \cap M_{\mathrm{mos}}$

        \If{$O_{r,c} = \emptyset$}
            \State $I_{\mathrm{mos}}(V_{r,c})
            \gets
            P_{r,c}(V_{r,c})$
        \Else
            \State $\Gamma^{\ast}
            \gets
            \textsc{DPSeam}(P_{r,c}, I_{\mathrm{mos}}, O_{r,c})$

            \State $\alpha
            \gets
            \textsc{FeatherMask}(\Gamma^{\ast})$

            \State $I_{\mathrm{mos}}
            \gets
            \textsc{AlphaBlend}(I_{\mathrm{mos}}, P_{r,c}, \alpha, V_{r,c})$
        \EndIf

        \State $M_{\mathrm{mos}}
        \gets
        M_{\mathrm{mos}} \cup V_{r,c}$
    \EndFor
\EndFor

\State $I_{\mathrm{out}} \gets I_{\mathrm{mos}}$

\Return $I_{\mathrm{out}}$

\end{algorithmic}
\end{algorithm}

Given an input large image and a user editing instruction,
the system first generates overlay tiles and binary masks, and then invokes a black-box VLM to generate multiple candidate results.
Each candidate is processed through geometric alignment, color-consistency correction, and seam-risk scoring.
Finally, the best candidate for each tile is fused back into the global canvas through dynamic-programming curved seam fusion.

\FloatBarrier

\section{Experiments}
\label{sec:experiments}

\subsection{Implementation Details}
\label{sec:impl}


SeamEdit is implemented in Python.
The core image-processing components rely on OpenCV and Pillow.
The VLM backend is invoked through either a web interface or an API.
The VLM backend used in the experiments is \textit{GPT-Image-2}.

The experimental configuration is shown in Table~\ref{tab:implementation-config}.
Since black-box VLM outputs are stochastic, the same tile can be generated multiple times to obtain multiple candidates.
SeamEdit automatically ranks candidate quality and selects the best available candidate for each tile before the final fusion stage.

In candidate ranking, the scoring weights in Eq.~\eqref{eq:candidate-score} are set to $\alpha_{\mathrm{psnr}}=1.2$, $\alpha_{\mathrm{ssim}}=20.0$, $\alpha_{\mathrm{seam}}=0.9$, $\alpha_{\mathrm{geo}}=1.0$, $\alpha_{\mathrm{valid}}=1.0$, and $\alpha_{\mathrm{color}}=1.2$.
The composite geometric penalty is expanded as $6.0E_{\mathrm{rmse}}+0.05|d_x|+0.05|d_y|+5.0|\theta|+180.0E_{\mathrm{scale}}+30.0\max(0,0.75-r_{\mathrm{inlier}})$, and the composite validity penalty is expanded as $0.9E_{\mathrm{black}}+0.6E_{\mathrm{crop}}$.
The color blending strength in Eq.~\eqref{eq:color-match-blend} is set to $\eta=0.65$, and the low-frequency color compensation strength in Eq.~\eqref{eq:lowfreq-color-correction} is set to $\gamma=0.75$.

\begin{table}[t]
\centering
\caption{Implementation configuration.}
\label{tab:implementation-config}
\begin{tabular}{lc}
\toprule
Setting & Value \\
\midrule
Input size & 1672 $\times$ 941 \\
Grid & 3 $\times$ 4 \\
Tiles & 12 \\
Overlay & 30 px / 120 px output \\
Scale & $\times$4 \\
Feature grid & 5 $\times$ 5 \\
SIFT features & 12000 \\
RANSAC threshold & 6 px \\
Feather width & 96 px \\
Color match strength & 0.65 \\
Low-frequency strength & 0.75 \\
Seam method & dynamic programming curved seam + color match + smoothed feather \\
Backend & GPT-Image-2 \\
\bottomrule
\end{tabular}
\end{table}

\subsection{Comparisons}
\label{sec:baselines}
\begin{figure}[t]
    \centering
    \includegraphics[width=\linewidth]{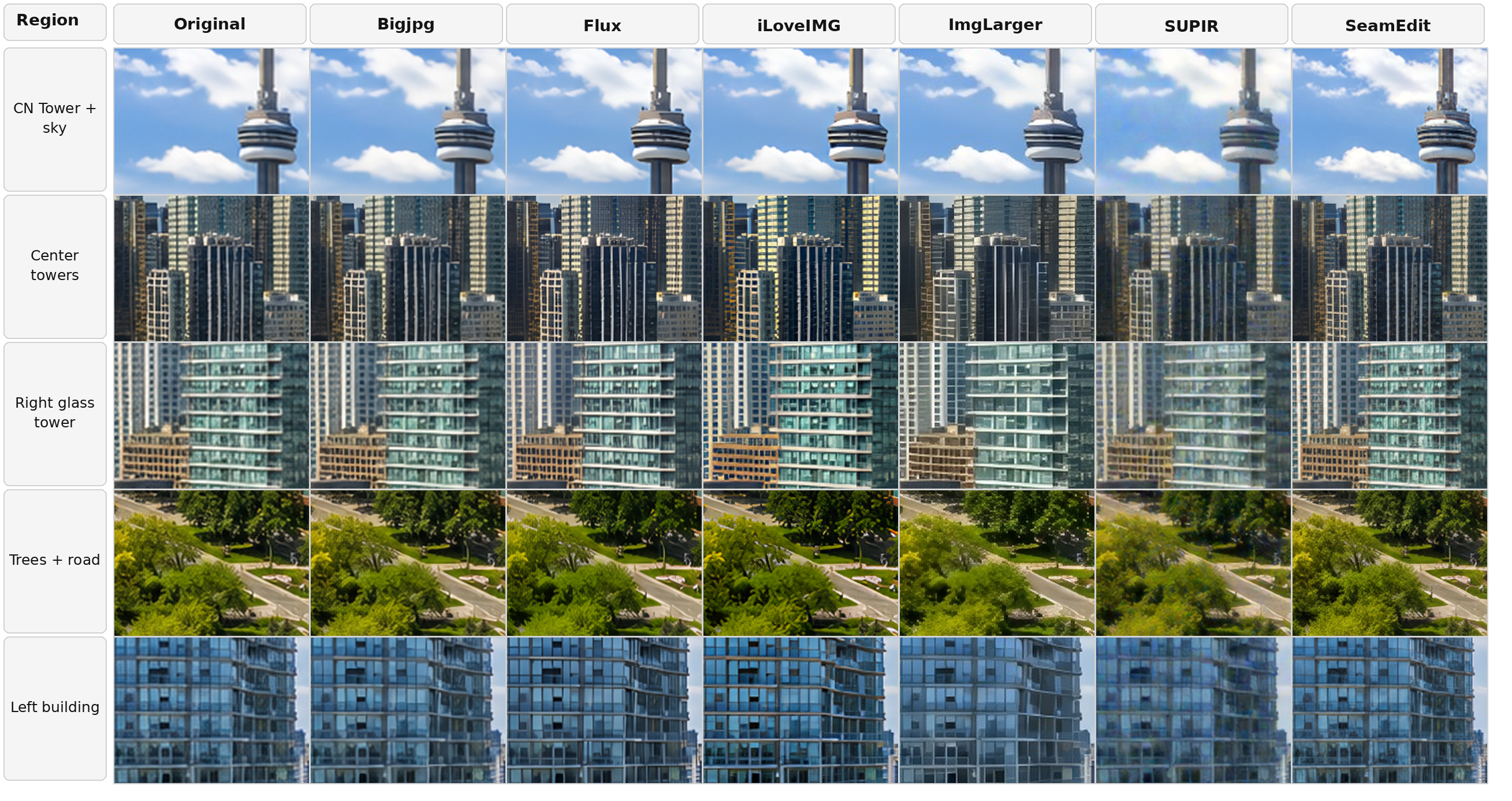}
    \caption{
    Zoomed qualitative comparison across different methods on representative local regions.
    Each row corresponds to the same image region, and each column corresponds to one method.
    The compared regions cover sky, towers, glass buildings, vegetation, roads, and dense building textures.
    SeamEdit maintains good visual quality in building structure, local texture, and overall color consistency.
    }
    \label{fig:qualitative_zoomed_comparison}
\end{figure}

We compare SeamEdit with two research baselines in qualitative experiments, and additionally include
commonly used online image enhancement tools as visual references:
\begin{itemize}
    \item \textbf{FLUX.2-Dev}~\cite{flux-2-2025}:
    an open-source generative image editing or repainting model used to compare structure preservation, texture consistency, and color continuity after local editing of large images.
    \item \textbf{SUPIR}~\cite{yu2024scaling}:
    an open-source image restoration and super-resolution method used to compare detail recovery, texture enhancement, and overall visual quality.
    \item \textbf{Bigjpg / iLoveIMG / ImgLarger}:
    online image enhancement tools used as references in qualitative comparison to observe visual differences in local structure, texture, and color consistency.
\end{itemize}

Figure~\ref{fig:qualitative_zoomed_comparison} shows zoomed qualitative comparisons across different methods on representative local regions.
Each row corresponds to the same image region, and each column corresponds to one method.
The compared regions cover sky, towers, glass buildings, vegetation, roads, and dense building textures.

\subsection{Qualitative Results}
\label{sec:quant}

Table~\ref{tab:candidate-ranking} reports quality statistics before and after automatic multi-candidate ranking.
``All candidates'' denotes all generated candidates, while ``Auto-selected'' denotes the final selected candidate for each tile.
In this case study, automatic ranking improves color difference, seam MAE, PSNR, and SSIM.
\begin{table}[t]
\centering
\caption{Automatic candidate selection quality.}
\label{tab:candidate-ranking}
\resizebox{\linewidth}{!}{%
\begin{tabular}{lccccccc}
\toprule
Group & $N$ & Score$\uparrow$ & RMSE$\downarrow$ & Color $\Delta\downarrow$ & Max Seam MAE$\downarrow$ & PSNR$\uparrow$ & SSIM$\uparrow$ \\
\midrule
All usable candidates & 15 & 26.54$\pm$26.35 & 2.25$\pm$0.53 & 24.09$\pm$28.84 & 67.23$\pm$66.66 & 17.30$\pm$5.89 & 0.723$\pm$0.336 \\
Auto-selected & 12 & 33.18$\pm$25.37 & 2.24$\pm$0.58 & 10.38$\pm$5.66 & 36.20$\pm$20.11 & 19.98$\pm$2.27 & 0.884$\pm$0.058 \\
\bottomrule
\end{tabular}%
}

\vspace{0.3em}
\footnotesize
\raggedright
Note: \(N\) denotes the number of usable candidates in this experiment. For generated results rejected by the model or below the minimum acceptance threshold of the system, new candidates were regenerated and added to the candidate pool. The final usable candidate pool contains 15 candidates. The Auto-selected row reports the candidates selected by automatic ranking for the 12 tiles.
\end{table}

This paper uses seam MAE to measure pixel-level discontinuity in boundary regions,
SSIM~\cite{wang2004ssim} to measure structural similarity,
and PSNR to measure pixel-level fidelity.
LPIPS~\cite{zhang2018lpips} can be used as a perceptual distance metric in extended baseline experiments.

\subsection{Module Effect Analysis}
\label{sec:ablation}

Table~\ref{tab:candidate-ranking} verifies the effect of multi-candidate quality ranking at the candidate level.
When a black-box VLM repeatedly generates the same tile, different candidates may exhibit different degrees of geometric drift, color deviation, and boundary artifacts.
Through geometric alignment, color-consistency correction, and seam-risk scoring, SeamEdit maps multiple candidates of each tile into a unified reference coordinate system for comparison,
and selects the best available candidate for the final fusion stage.

In the final fusion stage, the dynamic-programming curved seam preferentially passes through low-cost regions in the overlap band,
and combines color matching with smooth feathering to reduce boundary discontinuities.
Together, these modules reduce seam visibility after tiled editing.


\section{Discussion}
\label{sec:discussion}

\subsection{Limitations}

\textbf{Dependence on black-box backends.}
SeamEdit treats the VLM as a black-box API, and its generation quality changes with the capability of the selected model.
API changes or service discontinuation may affect reproducibility.
However, the pipeline design in this paper is decoupled from any specific backend, so switching to another VLM, including an open-source model, does not require modifications to the pipeline components.

\textbf{Empirical composite-score weights.}
The weights $\alpha_i, \beta_i$ in Eq.~\eqref{eq:candidate-score} are determined through empirical tuning.
A systematic sensitivity analysis of these empirical weights is left for future work.

\textbf{Multi-candidate call overhead.}
Generating $K$ candidates for each tile requires $K$ API calls, and the processing time scales linearly with $K$.
For time-sensitive applications, $K$ can be reduced, or multi-candidate generation can be applied only to tiles with high seam risk.

\textbf{Context conflict under extreme semantic edits.}
When the semantic modification of a tile differs substantially from its surrounding regions, the local context provided by the overlap area may conflict with the new semantics in the core region.
In this case, geometric correction and seam fusion can reduce boundary artifacts, while final semantic consistency still depends on the generation quality of the black-box VLM and the user-specified editing region.

\textbf{Applicable backend capability range.}
SeamEdit targets VLM foundation models with basic semantic understanding, defect localization, and visual inpainting capability.
For models whose generated results are severely degraded or semantically invalid, the main source of error comes from the foundation model itself, and including such models in the evaluation would confound backend generation failure with the performance of the repair strategy.
Therefore, this paper uses GPT-Image-2 as the main black-box backend for the final experiments.
Appendix~\ref{sec:Nano_Banana_2} additionally reports a backend failure case with Google Nano Banana 2, showing that severely degraded backend outputs can dominate the observed failure modes.
This setting matches SeamEdit's intended use case: further repairing model outputs that are generally usable but contain local defects.
Evaluating SeamEdit on more closed-source and open-source models with corresponding capabilities is an important direction for future work.

\subsection{Application Scenarios}

SeamEdit's VLM-agnostic design allows it to scale with future improvements in closed-source or open-source VLMs.
It is applicable to the following practical scenarios:

\textbf{Game art asset production.}
For local style iteration of large game maps or scene textures, traditional workflows require artists to manually handle boundaries.
SeamEdit can automate part of the seam-processing workflow and reduce the manual burden of boundary retouching.

\textbf{Local semantic modification of concept art.}
Concept designers can freely replace local terrain types, material styles, or vegetation categories in a large full image without regenerating the entire image.

\textbf{High-resolution image enhancement.}
The high-quality texture generation capability of VLMs can be applied to high-resolution tiles,
while SeamEdit handles seam transitions between adjacent tiles, forming a training-free high-resolution image enhancement workflow.

\section{Conclusion}
\label{sec:conclusion}

This paper introduced SeamEdit, a training-free and VLM-agnostic pipeline for semantic region editing in large images.
It is designed to mitigate two common issues that arise when black-box VLMs are applied to tiled image editing: alignment drift and seam artifacts.
The pipeline performs end-to-end editing through three collaborative modules:
overlay-based tile decomposition provides boundary context and seam buffers,
grid-SIFT alignment correction and color-consistency correction handle local drift and tonal differences,
seam-risk-driven multi-candidate ranking selects candidate results,
and dynamic-programming curved seam fusion reduces boundary visibility.

SeamEdit natively supports semantic modifications of arbitrary tile regions and has practical value in scenarios such as game art asset production, concept-art iteration, and high-resolution image enhancement.

As closed-source VLMs continue to improve,
SeamEdit's black-box-compatible design allows it to connect to different backends and provides an extensible processing framework for large-image semantic editing.

\section*{Author Contributions}
\noindent
\begin{tabular}{@{}p{0.16\linewidth}p{0.80\linewidth}@{}}
Xiangyu Lyu: & Conceptualization, methodology, implementation, experiments, analysis, and writing. \\
Dan Lei: & Fine-arts guidance and visual-quality assessment criteria.
\end{tabular}

\noindent

\section*{Funding}
No external funding was received for this work.

\bibliographystyle{plain}
\bibliography{./references}

\clearpage
\appendix
\section{Additional Supplementary Results}

\subsection{Additional Full-Image Results}

\begin{figure}[H]
    \centering
    \includegraphics[
        width=\textwidth,
        height=0.82\textheight,
        keepaspectratio
    ]{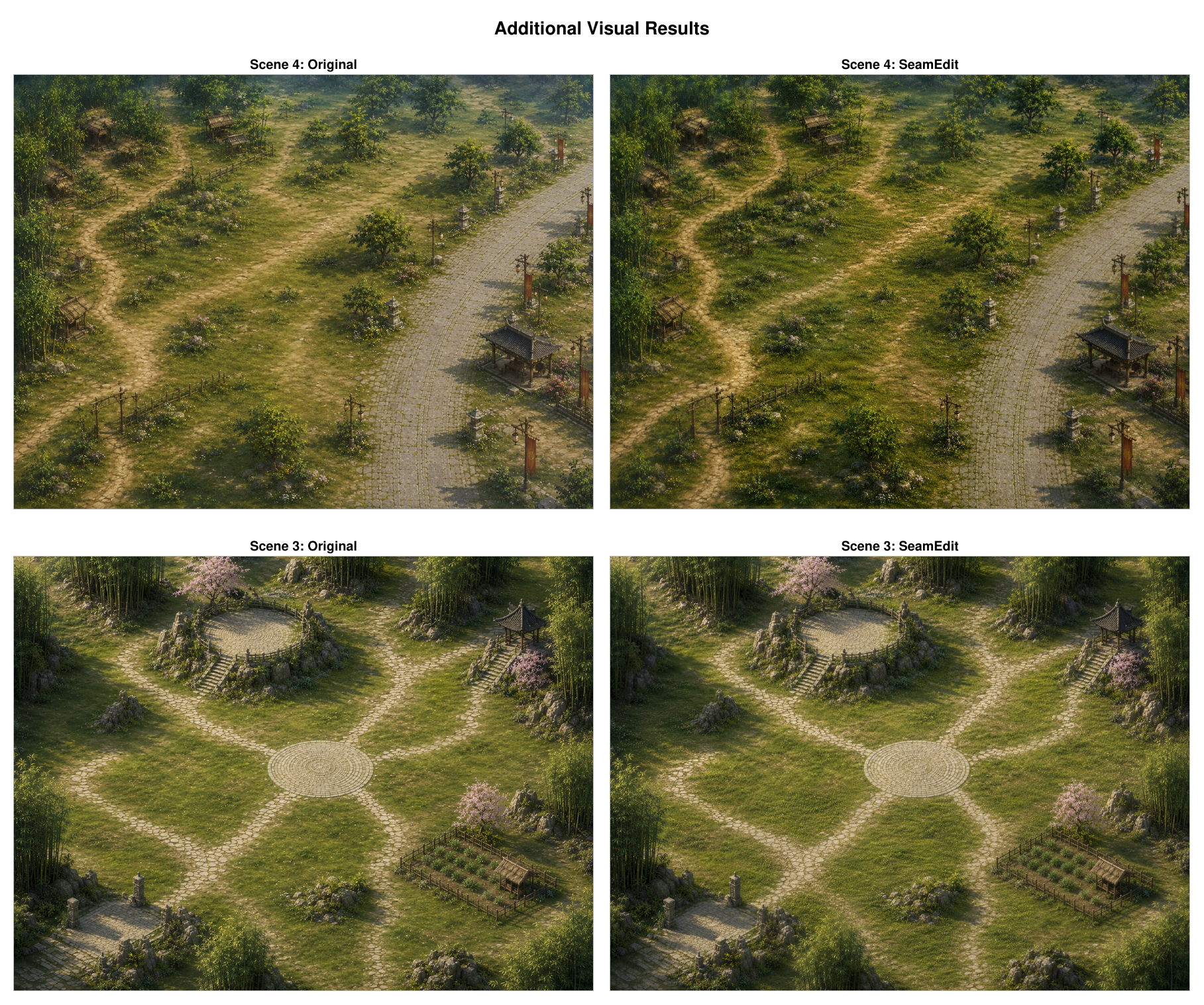}
    \caption{
    Additional full-image qualitative results. Each group shows the original image and the SeamEdit output.
    These results demonstrate SeamEdit's ability to preserve global layout, overall color tone, and semantic-region continuity at the full-image scale.
    }
    \label{fig:appendix-full-results}
\end{figure}

\subsection{Local Zoomed Qualitative Results}

\begin{figure}[H]
    \centering
    \includegraphics[
        width=\textwidth,
        height=0.82\textheight,
        keepaspectratio
    ]{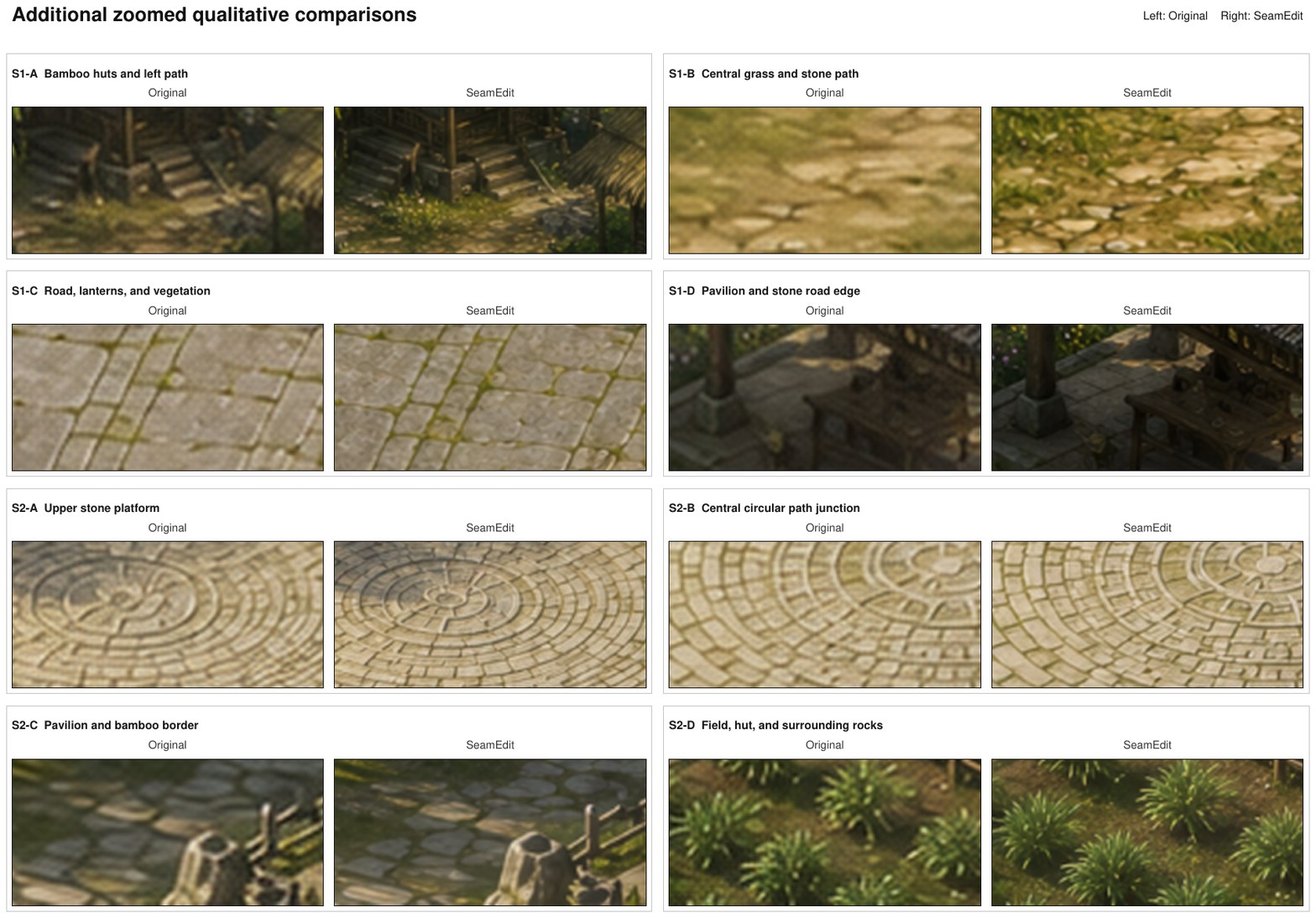}
    \caption{
    Additional local zoomed qualitative comparisons. Each subfigure compares the original image and the SeamEdit output over the same local region.
    The selected regions cover roads, vegetation, stone platforms, pavilions, bamboo boundaries, and local object edges,
    illustrating SeamEdit's effects on local texture, boundary transitions, and detail continuity.
    }
    \label{fig:appendix-zoomed-results}
\end{figure}

\clearpage
\subsection{Backend Failure Case: Nano Banana 2}\label{sec:Nano_Banana_2}

\begin{figure}[H]
    \centering
    \includegraphics[
        width=\textwidth,
        height=0.86\textheight,
        keepaspectratio
    ]{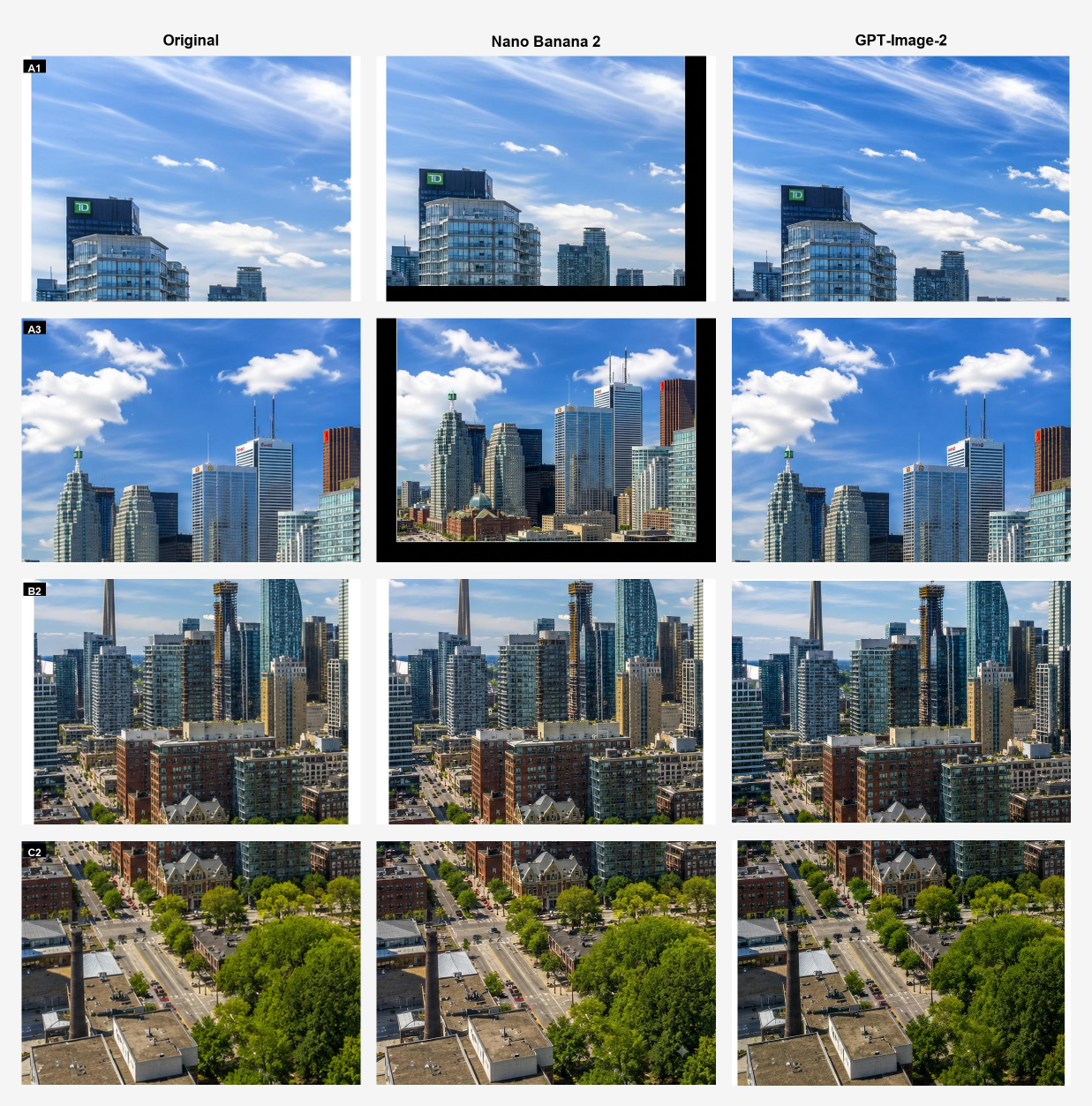}
    \caption{
    Qualitative comparison of different black-box VLM backends. The three columns show the original image, the Nano Banana 2 output, and the GPT-Image-2 output.
    Under the current large-image tiled editing setting, Nano Banana 2 tends to produce substantial structural rewriting, content hallucination, and weak correspondence to the input;
    GPT-Image-2 better preserves the original spatial layout and local structures. Therefore, the final experiments in this paper use GPT-Image-2 as the main black-box backend.
    }
    \label{fig:backend-failure-banana2}
\end{figure}

\end{document}